\documentclass[10pt,twocolumn,letterpaper]{article}

\usepackage{cvpr}
\usepackage{times}
\usepackage{epsfig}
\usepackage{graphicx}
\usepackage{amsmath}
\usepackage{amssymb}

\usepackage[pagebackref=true,breaklinks=true,letterpaper=true,colorlinks=true,citecolor=blue,bookmarks=false]{hyperref}
\usepackage[square,numbers,sort&compress]{natbib}
\usepackage{algpseudocode}
\usepackage{enumitem}
\usepackage{adjustbox}
\usepackage{booktabs}
\usepackage{array}
\usepackage{multirow}
\usepackage{algorithm}
\usepackage{caption}
\usepackage{subcaption}
\usepackage{color, colortbl}
\usepackage[normalem]{ulem}

\newcolumntype{L}[1]{>{\raggedright\let\newline\\\arraybackslash\hspace{0pt}}m{#1}}
\newcolumntype{C}[1]{>{\centering\let\newline\\\arraybackslash\hspace{0pt}}m{#1}}
\newcolumntype{R}[1]{>{\raggedleft\let\newline\\\arraybackslash\hspace{0pt}}m{#1}}
\definecolor{Gray}{gray}{0.7}

\cvprfinalcopy 

\def\cvprPaperID{1604} 

\ifcvprfinal\fi
\begin{document}

\title{Cross-stitch Networks for Multi-task Learning}

\author{
    Ishan Misra\thanks{Both authors contributed equally} \quad \quad Abhinav Shrivastava\footnotemark[1] \quad \quad Abhinav Gupta \quad \quad Martial Hebert \\
    The Robotics Institute, Carnegie Mellon University
}

\maketitle

\begin{abstract}
Multi-task learning in Convolutional Networks has displayed remarkable success in the field of recognition. This success can be largely attributed to learning shared representations from multiple supervisory tasks. However, existing multi-task approaches rely on enumerating multiple network architectures specific to the tasks at hand, that do not generalize. In this paper, we propose a principled approach to learn shared representations in ConvNets using multi-task learning. Specifically, we propose a new sharing unit: ``cross-stitch'' unit. These units combine the activations from multiple networks and can be trained end-to-end. A network with cross-stitch units can learn an optimal combination of shared and task-specific representations. Our proposed method generalizes across multiple tasks and shows dramatically improved performance over baseline methods for categories with few training examples.
\end{abstract}

\section{Introduction}
\begin{figure}[t]
\centering 
\includegraphics[width=0.48\textwidth]{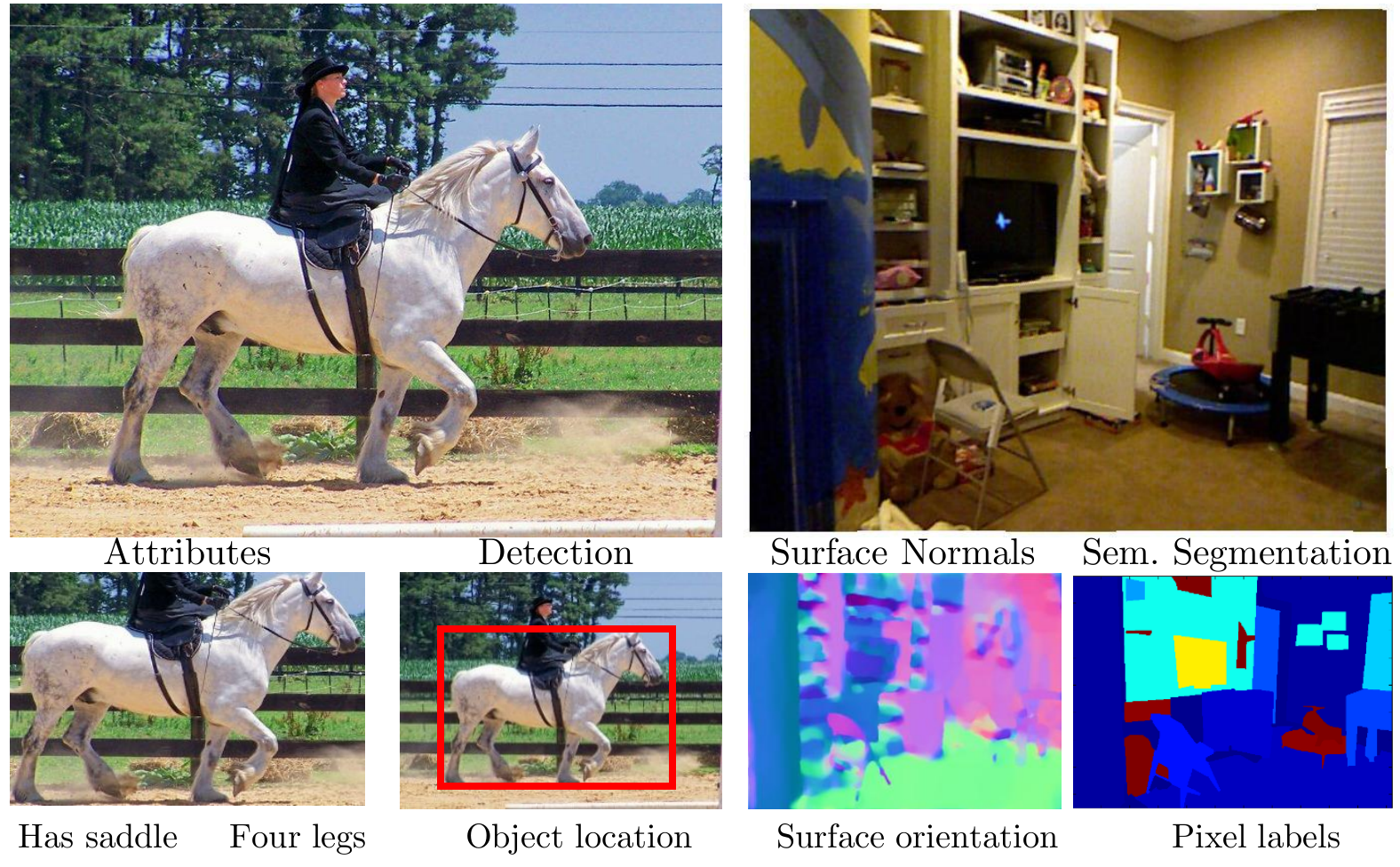}
\caption{Given an input image, one can leverage multiple related properties to improve performance by using a multi-task learning framework. In this paper, we propose \emph{cross-stitch} units, a principled way to use such a multi-task framework for ConvNets.}
\label{fig:teaser}
\end{figure}

Over the last few years, ConvNets have given huge performance boosts in recognition tasks ranging from classification and detection to segmentation and even surface normal estimation.
One of the reasons for this success is attributed to the inbuilt sharing mechanism, which allows ConvNets to learn representations shared across different categories. 
This insight naturally extends to sharing between tasks (see Figure~\ref{fig:teaser}) and leads to further performance improvements, \eg, the gains in segmentation~\cite{SDS} and detection~\cite{fast-rcnn,nikos}.
A key takeaway from these works is that multiple tasks, and thus multiple types of supervision, helps achieve better performance with the same input. 
But unfortunately, the network architectures used by them for multi-task learning notably differ. There are no insights or principles for how one should choose ConvNet architectures for multi-task learning.

\subsection{Multi-task sharing: an empirical study}
\label{sec:intro-study}
How should one pick the right architecture for multi-task learning? Does it depend on the final tasks? Should we have a completely shared representation between tasks? Or should we have a combination of shared and task-specific representations? Is there a principled way of answering these questions?

\begin{figure*}[t]
\centering
\includegraphics[width=\textwidth]{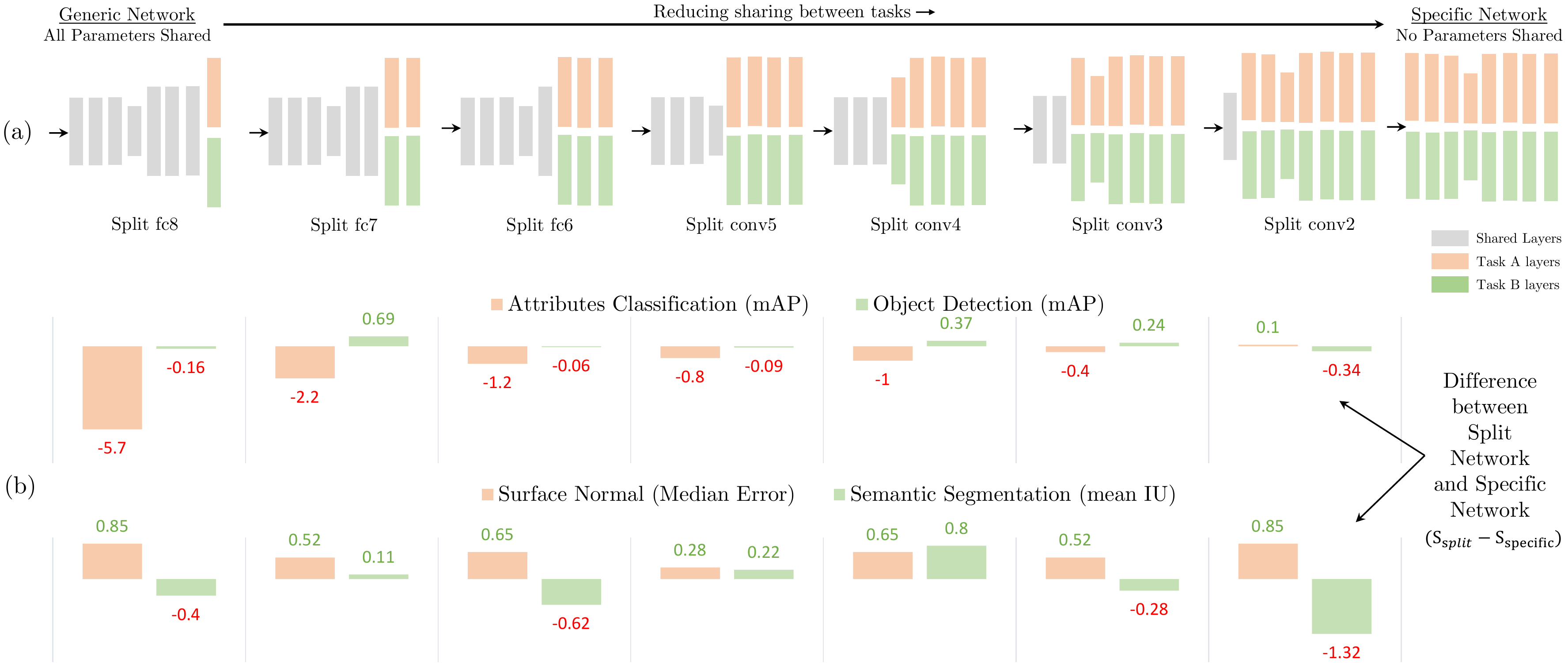}
\caption{We train a variety of multi-task (two-task) architectures by splitting at different layers in a ConvNet~\cite{alexnet} for two pairs of tasks. For each of these networks, we plot their performance on each task relative to the task-specific network. We notice that the best performing multi-task architecture depends on the individual tasks and does not transfer across different pairs of tasks.}
\label{fig:diffsplits}
\vspace{-0.1in}
\end{figure*}

To investigate these questions, we first perform extensive experimental analysis to understand the performance trade-offs amongst different combinations of shared and task-specific representations. Consider a simple experiment where we train a ConvNet on two related tasks (\eg, semantic segmentation and surface normal estimation). Depending on the amount of sharing one wants to enforce, there is a spectrum of possible network architectures. Figure~\ref{fig:diffsplits}(a) shows different ways of creating such network architectures based on AlexNet~\cite{alexnet}. On one end of the spectrum is a fully shared representation where all layers, from the first convolution (\texttt{conv2}) to the last fully-connected (\texttt{fc7}), are shared and only the last layers (two \texttt{fc8}s) are task specific. An example of such sharing is~\cite{fast-rcnn} where separate \texttt{fc8} layers are used for classification and bounding box regression. On the other end of the sharing spectrum, we can train two networks separately for each task and there is no cross-talk between them. In practice, different amount of sharing tends to work best for different tasks.

So given a pair of tasks, how should one pick a network architecture? To empirically study this question, we pick two varied pairs of tasks:

\begin{itemize}[leftmargin=*]
\itemsep0em
\item We first pair semantic segmentation (SemSeg) and surface normal prediction (SN). We believe the two tasks are closely related to each other since segmentation boundaries also correspond to surface normal boundaries. For this pair of tasks, we use NYU-v2~\cite{nyuv2} dataset.

\item For our second pair of tasks we use detection (Det) and Attribute prediction (Attr). Again we believe that two tasks are related: for example, a box labeled as ``car'' would also be a positive example of ``has wheel'' attribute. For this experiment, we use the attribute PASCAL dataset~\cite{pascal,apascal}.
\end{itemize}

We exhaustively enumerate all the possible \emph{Split} architectures as shown in Figure~\ref{fig:diffsplits}(a) for these two pairs of tasks and  show their respective performance in Figure~\ref{fig:diffsplits}(b). The best performance for both the SemSeg and SN tasks is using the ``Split \texttt{conv4}'' architecture (splitting at \texttt{conv4}), while for the Det task it is using the Split \texttt{conv2}, and for Attr with Split \texttt{fc6}. These results indicate two things -- 1) Networks learned in a multi-task fashion have an edge over networks trained with one task; and 2) The best Split architecture for multi-task learning depends on the tasks at hand.

While the gain from multi-task learning is encouraging, getting the most out of it is still cumbersome in practice. This is largely due to the task dependent nature of picking architectures and the lack of a principled way of exploring them. Additionally, enumerating all possible architectures for each set of tasks is impractical. This paper proposes \emph{cross-stitch units}, using which a single network can capture all these Split-architectures (and more). It automatically learns an optimal combination of shared and task-specific representations. We demonstrate that such a \emph{cross-stitched} network can achieve better performance than the networks found by brute-force enumeration and search.
\section{Related Work}
Generic Multi-task learning~\cite{stein1956inadmissibility,caruanaThesis} has a rich history in machine learning. The term \emph{multi-task learning} (MTL) itself has been broadly used~\cite{jalali2010dirty,romera2012exploiting,amit2007uncovering,yang2014unified,xue2007multi,evgeniou2004regularized} as an umbrella term to include representation learning and selection~\cite{evgeniou2007multi,argyriou2008convex,whomToShare,obozinski2006multi}, transfer learning~\cite{pan2010survey,yosinski2014transferable, razavian2014cnn} \etc and their widespread applications in other fields, such as genomics~\cite{obozinski2010joint}, natural language processing~\cite{collobert2011natural,collobert2008unified,liu2015representation} and computer vision~\cite{whomToShare,zhang2013robust, wright2010sparse,donahue2013decaf,jung2015rotating,torralba2007sharing,quattoni2008transfer,pentina2015curriculum}. In fact, many times multi-task learning is implicitly used without reference; a good example being fine-tuning or transfer learning~\cite{razavian2014cnn}, now a mainstay in computer vision, can be viewed as sequential multi-task learning~\cite{caruanaThesis}. Given the broad scope, in this section we focus only on multi-task learning in the context of ConvNets used in computer vision.

Multi-task learning is generally used with ConvNets in computer vision to model related tasks jointly, \eg pose estimation and action recognition~\cite{gkioxari2014r}, surface normals and edge labels~\cite{wangfouhey}, face landmark detection and face detection~\cite{zhang2014facial, zhang2014improving}, auxiliary tasks in detection~\cite{fast-rcnn}, related classes for image classification~\cite{teterwakshared} \etc. Usually these methods share some features (layers in ConvNets) amongst tasks and have some task-specific features. This sharing or split-architecture (as explained in Section~\ref{sec:intro-study}) is decided after experimenting with splits at multiple layers and picking the best one. Of course, depending on the task at hand, a different Split architecture tends to work best, and thus given new tasks, new split architectures need to be explored. In this paper, we propose \emph{cross-stitch units} as a principled approach to explore and embody such Split architectures, without having to train all of them.

In order to demonstrate the robustness and effectiveness of cross-stitch units in multi-task learning, we choose varied tasks on multiple datasets. In particular, we select four well established and diverse tasks on different types of image datasets: 1) We pair semantic segmentation~\cite{shi2000normalized,shotton2006textonboost,heitz2008learning} and surface normal estimation~\cite{3dp,eigen2015predicting,wangfouhey}, both of which require predictions over all pixels, on the NYU-v2 indoor dataset~\cite{nyuv2}. These two tasks capture both semantic and geometric information about the scene. 2) We choose the task of object detection~\cite{schneiderman2004object,pedro-dpm,girshick2014rich,fast-rcnn} and attribute prediction~\cite{farhadi2010attribute,abdulnabi2015multi,lampert2009learning} on web-images from the PASCAL dataset~\cite{apascal,pascal}. These tasks make predictions about localized regions of an image.
\begin{figure}[t]
\centering
\includegraphics[width=0.4\textwidth]{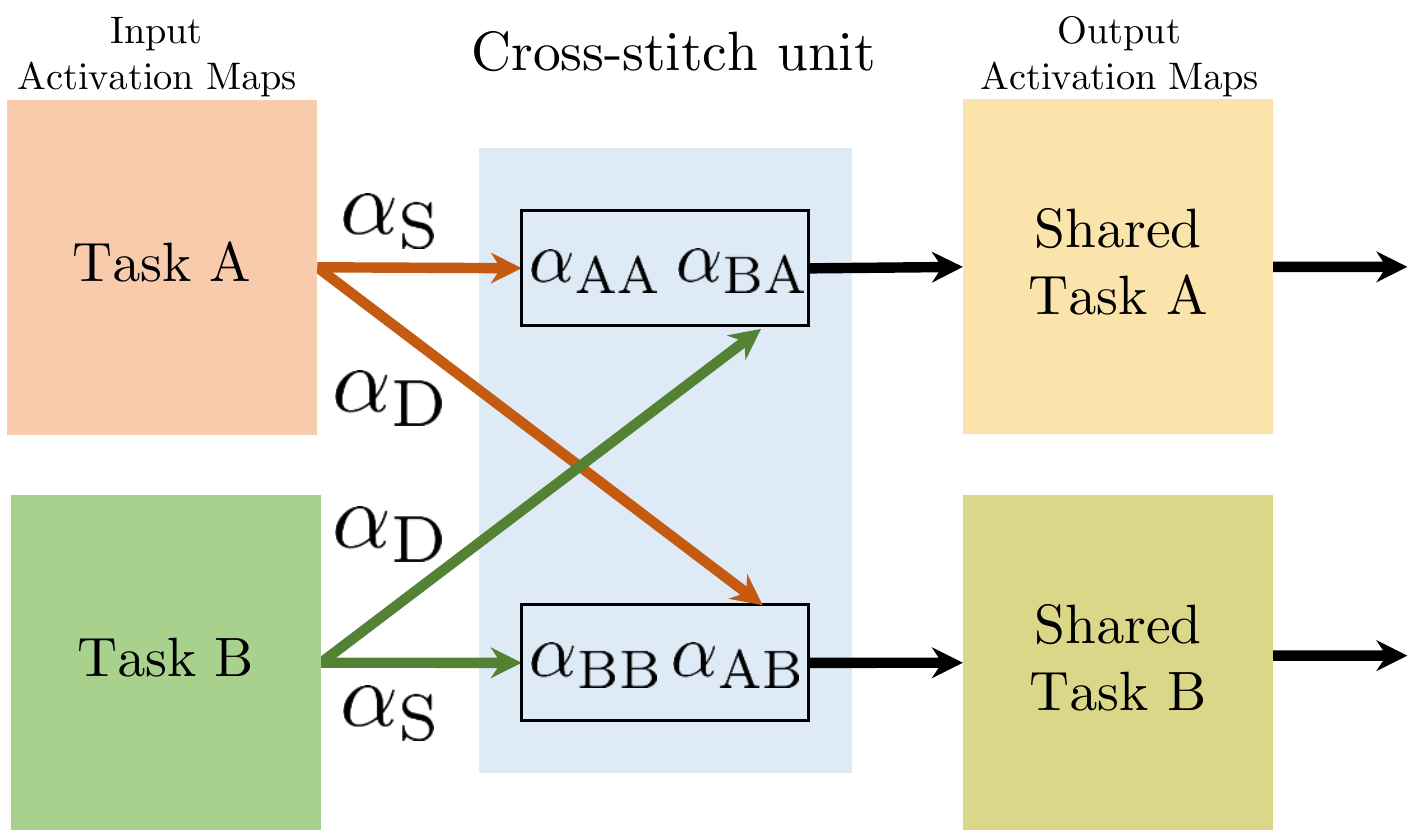}
\caption{We model shared representations by learning a linear combination of input activation maps. At each layer of the network, we learn such a linear combination of the activation maps from both the tasks. The next layers' filters operate on this shared representation.}
\vspace{-0.1in}
\label{fig:lincomb-detail}
\end{figure}

\section{Cross-stitch Networks}
In this paper, we present a novel approach to multi-task learning for ConvNets by proposing cross-stitch units. Cross-stitch units try to find the best shared representations for multi-task learning. They model these shared representations using linear combinations, and learn the optimal linear combinations for a given set of tasks. We integrate these cross-stitch units into a ConvNet and provide an end-to-end learning framework. We use detailed ablative studies to better understand these units and their training procedure. Further, we demonstrate the effectiveness of these units for two different pairs of tasks. To limit the scope of this paper, we only consider tasks which take the same single input, \eg, an image as opposed to say an image and a depth-map~\cite{rcnn-depth}.

\subsection{Split Architectures}
Given a single input image with multiple labels, one can design ``Split architectures'' as shown in Figure~\ref{fig:diffsplits}. These architectures have both a shared representation and a task specific representation. `Splitting' a network at a lower layer allows for more task-specific and fewer shared layers. One extreme of Split architectures is splitting at the lowest convolution layer which results in two separate networks altogether, and thus only task-specific representations. The other extreme is using ``sibling'' prediction layers (as in~\cite{fast-rcnn}), which allows for a more shared representation. Thus, Split architectures allow for a varying amount of shared and task-specific representations.

\subsection{Unifying Split Architectures}
Given that Split architectures hold promise for multi-task learning, an obvious question is -- At which layer of the network should one split? This decision is highly dependent on the input data and tasks at hand. Rather than enumerating the possibilities of Split architectures for every new input task, we propose a simple architecture that can learn how much shared and task specific representation to use.

\subsection{Cross-stitch units}
Consider a case of multi task learning with two tasks $\mathrm{A}$ and $\mathrm{B}$ on the same input image. For the sake of explanation, consider two networks that have been trained separately for these tasks. We propose a new unit, \emph{cross-stitch unit}, that combines these two networks into a multi-task network in a way such that the tasks supervise how much sharing is needed, as illustrated in Figure~\ref{fig:lincomb-detail}. At each layer of the network, we model sharing of representations by learning a linear combination of the activation maps~\cite{whomToShare,argyriou2008convex} using a \emph{cross-stitch unit}. Given two activation maps $x_\mathrm{A}, x_\mathrm{B}$ from layer $l$ for both the tasks, we learn linear combinations $\tilde{x}_\mathrm{A}, \tilde{x}_\mathrm{B}$ (Eq~\ref{eqn:forward}) of both the input activations and feed these combinations as input to the next layers' filters. This linear combination is parameterized using $\alpha$. Specifically, at location $\left(i,j\right)$ in the activation map,
\begin{align}
\label{eqn:forward}
\begin{bmatrix}
\tilde{x}^{ij}_\mathrm{A} \\[2pt]
\tilde{x}^{\strut ij}_\mathrm{B}
\end{bmatrix}
&= \begin{bmatrix}
\alpha_\mathrm{AA}  & \alpha_\mathrm{AB} \\
\alpha_{BA}  & \alpha_\mathrm{BB} \\
\end{bmatrix}
\begin{bmatrix}
x^{ij}_\mathrm{A} \\[2pt]
x^{\strut ij}_\mathrm{B}
\end{bmatrix}
 \end{align}
We refer to this the \emph{cross-stitch} operation, and the unit that models it for each layer $l$ as the \emph{cross-stitch unit}. The network can decide to make certain layers task specific by setting $\alpha_\mathrm{AB}$ or $\alpha_\mathrm{BA}$ to zero, or choose a more shared representation by assigning a higher value to them.

\paragraph{Backpropagating through cross-stitch units.} Since cross-stitch units are modeled as linear combination, their partial derivatives for loss $L$ with tasks $\mathrm{A}, \mathrm{B}$ are computed as

\begin{align}
\label{eqn:backward}
\begin{bmatrix}
\frac{\strut\partial L}{\strut\partial x^{ij}_\mathrm{A}} \\
\frac{\strut\partial L}{\strut\partial x^{ij}_\mathrm{B}}
\end{bmatrix} &=
\begin{bmatrix}
\alpha_\mathrm{AA}  & \alpha_\mathrm{BA} \\
\alpha_\mathrm{AB}  & \alpha_\mathrm{BB} \\
\end{bmatrix}
\begin{bmatrix}
\frac{\strut\partial L}{\strut\partial \tilde{x}^{ij}_\mathrm{A}} \\
\frac{\strut\partial L}{\strut\partial \tilde{x}^{ij}_\mathrm{B}}
\end{bmatrix} \\
\frac{\strut\partial L}{\strut\partial{\alpha_\mathrm{AB}}} &= \frac{\strut\partial L}{\strut\partial \tilde{x}^{ij}_\mathrm{B}}  x^{ij}_\mathrm{A}, \;\;\;
\frac{\strut\partial L}{\strut\partial{\alpha_\mathrm{AA}}} = \frac{\strut\partial L}{\strut\partial \tilde{x}^{ij}_\mathrm{A}}  x^{ij}_\mathrm{A}
\end{align}

We denote $\alpha_\mathrm{AB}, \alpha_\mathrm{BA}$ by $\alpha_\mathrm{D}$ and call them the \emph{different}-task values because they weigh the activations of another task. Likewise, $\alpha_\mathrm{AA}, \alpha_\mathrm{BB}$ are denoted by $\alpha_\mathrm{S}$, the \emph{same}-task values, since they weigh the activations of the same task.
By varying $\alpha_\mathrm{D}$ and $\alpha_\mathrm{S}$ values, the unit can freely move between shared and task-specific representations, and choose a middle ground if needed.
\section{Design decisions for cross-stitching}
\label{sec:twoTowers}
\begin{figure}[!ht]
\centering
\includegraphics[width=0.5\textwidth]{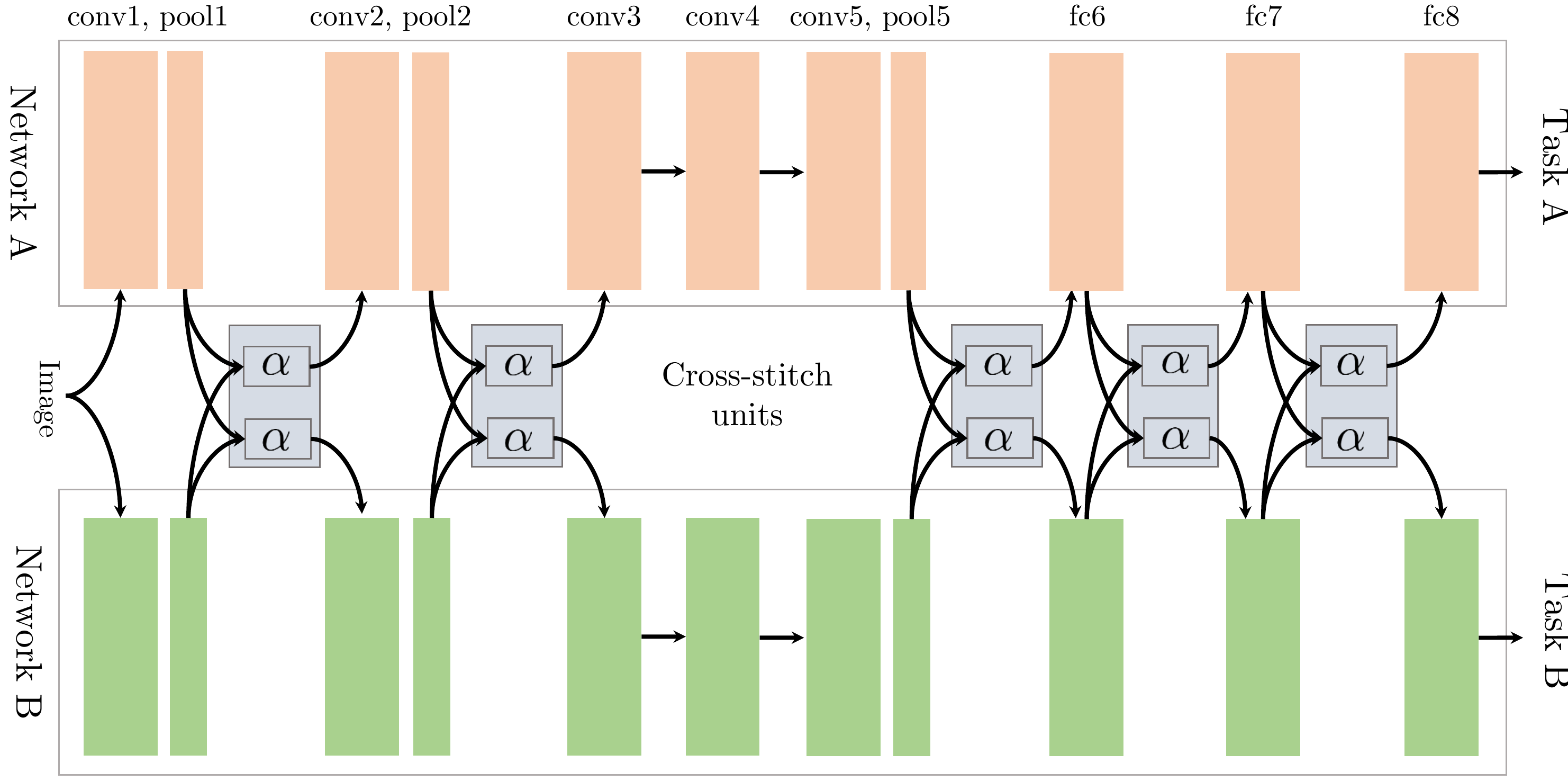}
\caption{Using cross-stitch units to stitch two AlexNet~\cite{alexnet} networks. In this case, we apply cross-stitch units only after pooling layers and fully connected layers. Cross-stitch units can model shared representations as a linear combination of input activation maps. This network tries to learn representations that can help with both tasks $\mathrm{A}$ and $\mathrm{B}$. We call the sub-network that gets direct supervision from task $\mathrm{A}$ as network $\mathrm{A}$ (top) and the other as network $\mathrm{B}$ (bottom).}
\label{fig:cross-stitch-two}
\end{figure}

We use the cross-stitch unit for multi-task learning in ConvNets. For the sake of simplicity, we assume multi-task learning with two tasks. Figure~\ref{fig:cross-stitch-two} shows this architecture for two tasks $\mathrm{A}$ and $\mathrm{B}$. The sub-network in Figure~\ref{fig:cross-stitch-two}(top) gets direct supervision from task $\mathrm{A}$ and indirect supervision (through cross-stitch units) from task $\mathrm{B}$. We call the sub-network that gets direct supervision from task $\mathrm{A}$ as network $\mathrm{A}$, and correspondingly the other as $\mathrm{B}$.
Cross-stitch units help regularize both tasks by learning and enforcing shared representations by combining activation (feature) maps.
As we show in our experiments, in the case where one task has less labels than the other, such regularization helps the ``data-starved'' tasks. 

Next, we enumerate the design decisions when using cross-stitch units with networks, and in later sections perform ablative studies on each of them.

\par \noindent \textbf{Cross-stitch units initialization and learning rates:} The $\alpha$ values of a cross-stitch unit model linear combinations of feature maps. Their initialization in the range $\left[0,1\right]$ is important for stable learning, as it ensures that values in the output activation map (after cross-stitch unit) are of the same order of magnitude as the input values before linear combination. We study the impact of different initializations and learning rates for cross-stitch units in Section~\ref{sec:ablation}.

\par \noindent \textbf{Network initialization:} Cross-stitch units combine together two networks as shown in Figure~\ref{fig:cross-stitch-two}. However, an obvious question is -- how should one initialize the networks $\mathrm{A}$ and $\mathrm{B}$? We can initialize networks $\mathrm{A}$ and $\mathrm{B}$ by networks that were trained on these tasks separately, or have the same initialization and train them jointly.

\section{Ablative analysis}
\label{sec:ablation}
We now describe the experimental setup in detail, which is common throughout the ablation studies.

\par \noindent \textbf{Datasets and Tasks:} For ablative analysis we consider the tasks of semantic segmentation (SemSeg) and Surface Normal Prediction (SN) on the NYU-v2~\cite{nyuv2} dataset. We use the standard train/test splits from~\cite{3dp}. For semantic segmentation, we follow the setup from~\cite{gupta2013} and evaluate on the 40 classes using the standard metrics from their work

\par \noindent \textbf{Setup for Surface Normal Prediction:}
Following~\cite{wangfouhey}, we cast the problem of surface normal prediction as classification into one of 20 categories. 
For evaluation, we convert the model predictions to 3D surface normals and apply the Manhattan-World post-processing following the method in~\cite{wangfouhey}.
We evaluate all our methods using the metrics from~\cite{3dp}. These metrics measure the error in the ground truth normals and the predicted normals in terms of their angular distance (measured in degrees). Specifically, they measure the mean and median error in angular distance, in which case lower error is better (denoted by `Mean' and `Median' error). They also report percentage of pixels which have their angular distance under a threshold (denoted by `Within $t^{\circ}$' at a threshold of $11.25^{\circ}, 22.5^{\circ}, 30^{\circ}$), in which case a higher number indicates better performance.

\par \noindent \textbf{Networks:} For semantic segmentation (SemSeg) and surface normal (SN) prediction, we use the Fully-Convolutional Network (FCN 32-s) architecture from~\cite{fcn} based on CaffeNet~\cite{caffe} (essentially AlexNet~\cite{alexnet}). For both the tasks of SemSeg and SN, we use RGB images at full resolution, and use mirroring and color data augmentation.
We then finetune the network (referred to as \emph{one-task network}) from ImageNet~\cite{imagenet} for each task using hyperparameters reported in~\cite{fcn}. We fine-tune the network for semantic segmentation for 25k iterations using SGD (mini-batch size 20) and for surface normal prediction for 15k iterations (mini-batch size 20) as they gave the best performance, and further training (up to 40k iterations) showed no improvement. These \emph{one-task networks} serve as our baselines and initializations for cross-stitching, when applicable.

\par \noindent \textbf{Cross-stitching:} We combine two AlexNet architectures using the cross-stitch units as shown in Figure~\ref{fig:cross-stitch-two}. We experimented with applying cross-stitch units after every convolution activation map and after every pooling activation map, and found the latter performed better. Thus, the cross-stitch units for AlexNet are applied on the activation maps for \texttt{pool1}, \texttt{pool2}, \texttt{pool5}, \texttt{fc6} and \texttt{fc7}. We maintain one cross-stitch unit per `channel' of the activation map, \eg, for \texttt{pool1} we have 96 cross-stitch units.

\begin{table}[t]
	\setlength{\tabcolsep}{0.2em}
	\centering
	\footnotesize{
		\caption{Initializing cross-stitch units with different $\alpha$ values, each corresponding to a convex combination. Higher values for $\alpha_\mathrm{S}$ indicate that we bias the cross-stitch unit to prefer task specific representations. The cross-stitched network is robust across different initializations of the units.}
		\label{tbl:init-cross-stitch}
		\resizebox{\linewidth}{!}{
			\begin{tabular}{@{}L{1.2cm} r*{7}{C{0.75cm}}@{}}
				\toprule
				& \multicolumn{5}{c}{{ Surface Normal}} & \multicolumn{3}{c}{{ Segmentation}} \\
				\arrayrulecolor{Gray}
				\cmidrule(l{-0.1cm}r{1.5pt}){2-6}
				\cmidrule(l{1.5pt}){7-9}
				\arrayrulecolor{black}
				& \multicolumn{2}{>{\centering\let\newline\\\arraybackslash\hspace{-0.15cm}}m{1.59cm}}{Angle Distance} & \multicolumn{3}{c}{Within $t^{\circ} $} & & & \\
				& \multicolumn{2}{>{\centering\let\newline\\\arraybackslash\hspace{-0.15cm}}m{1.59cm}}{(Lower Better)} & \multicolumn{3}{c}{(Higher Better)} & \multicolumn{3}{c}{(Higher Better)}\\
				$\left(\alpha_\mathrm{S}, \alpha_\mathrm{D}\right)$  & Mean & Med.\ & 11.25 & 22.5 & 30 & pixacc & mIU & fwIU \\
				\midrule
				$\left(0.1, 0.9\right)$ & 34.6 & 18.8 & 38.5 & 53.7 & 59.4 & 47.9 & 18.2 & 33.3 \\
				$\left(0.5, 0.5\right)$ & 34.4 & 18.8 & 38.5  & 53.7 & 59.5 & 47.2 & 18.6 & 33.8 \\
				$\left(0.7, 0.3\right)$ & \textbf{34.0} & \textbf{18.3} & 38.9 &	54.3 & 60.1 & 48.0	& 18.6 & 33.6 \\
				$\left(0.9, 0.1\right)$ & \textbf{34.0} &	\textbf{18.3} &	\textbf{39.0} & \textbf{54.4} & \textbf{60.2} & \textbf{48.2} & \textbf{18.9} & \textbf{34.0}\\
				\bottomrule
			\end{tabular}
		}
	}
\end{table}

\subsection{Initializing parameters of cross-stitch units}
\label{sec:init-unit}

Cross-stitch units capture the intuition that shared representations can be modeled by linear combinations~\cite{whomToShare}. 
To ensure that values after the cross-stitch operation are of the same order of magnitude as the input values, an obvious initialization of the unit is that the $\alpha$ values form a convex linear combination, \ie, the different-task $\alpha_\mathrm{D}$ and the same-task $\alpha_\mathrm{S}$ to sum to one. Note that this convexity is not enforced on the $\alpha$ values in either Equation~\ref{eqn:forward} or~\ref{eqn:backward}, but serves as a reasonable initialization. For this experiment, we initialize the networks $\mathrm{A}$ and $\mathrm{B}$ with \emph{one-task} networks that were fine-tuned on the respective tasks. Table~\ref{tbl:init-cross-stitch} shows the results of evaluating cross-stitch networks for different initializations of $\alpha$ values.

\subsection{Learning rates for cross-stitch units}
We initialize the $\alpha$ values of the cross-stitch units in the range $[0.1,0.9]$, which is about one to two orders of magnitude larger than the typical range of layer parameters in AlexNet~\cite{alexnet}. While training, we found that the gradient updates at various layers had magnitudes which were reasonable for updating the layer parameters, but too small for the cross-stitch units. Thus, we use higher learning rates for the cross-stitch units than the base network. In practice, this leads to faster convergence and better performance. To study the impact of different learning rates, we again use a cross-stitched network initialized with two \emph{one-task networks}. We scale the learning rates (wrt.\ the network's learning rate) of cross-stitch units in powers of $10$ (by setting the \texttt{lr\_mult} layer parameter in Caffe~\cite{caffe}). Table~\ref{tbl:lr-init} shows the results of using different learning rates for the cross-stitch units after training for 10k iterations. Setting a higher scale for the learning rate improves performance, with the best range for the scale being $10^2 - 10^3$. We observed that setting the scale to an even higher value made the loss diverge.

\begin{table}[t]
\setlength{\tabcolsep}{0.3em}
\centering
\footnotesize{
\caption{Scaling the learning rate of cross-stitch units wrt.\ the base network. Since the cross-stitch units are initialized in a different range from the layer parameters, we scale their learning rate for better training.}
\label{tbl:lr-init}
\resizebox{\linewidth}{!}{
\begin{tabular}{@{}L{0.8cm} r*{7}{C{0.75cm}}@{}}
\toprule
& \multicolumn{5}{c}{{ Surface Normal}} & \multicolumn{3}{c}{{ Segmentation}} \\
\arrayrulecolor{Gray}
\cmidrule(l{-0.1cm}r{1.5pt}){2-6}
\cmidrule(l{1.5pt}){7-9}
\arrayrulecolor{black}
& \multicolumn{2}{>{\centering\let\newline\\\arraybackslash\hspace{-0.15cm}}m{1.59cm}}{Angle Distance} & \multicolumn{3}{c}{Within $t^{\circ} $} & & & \\

& \multicolumn{2}{>{\centering\let\newline\\\arraybackslash\hspace{-0.15cm}}m{1.59cm}}{(Lower Better)} & \multicolumn{3}{c}{(Higher Better)} & \multicolumn{3}{c}{(Higher Better)}\\
Scale & Mean & Med. & 11.25 & 22.5 & 30 & pixacc & mIU & fwIU \\
\midrule
 1 & 34.6 & 18.9 & 38.4 & 53.7 & 59.4 & 47.7 & 18.6 & 33.5\\
 10 & 34.5 & 18.8 & 38.5 & 53.8 & 59.5 & 47.8 & 18.7 & 33.5\\
 $10^2$ & \textbf{34.0} & 18.3 & 39.0 & 54.4 & 60.2 & \textbf{48.0} & 18.9 & 33.8 \\
 $10^3$ & 34.1 & \textbf{18.2} & \textbf{39.2} & \textbf{54.4} & \textbf{60.2} & 47.2 & \textbf{19.3} & \textbf{34.0} \\
\bottomrule
\end{tabular}
}
}
\end{table}

\subsection{Initialization of networks $\mathrm{A}$ and $\mathrm{B}$}
When cross-stitching two networks, how should one initialize the networks $\mathrm{A}$ and $\mathrm{B}$? Should one start with task specific \emph{one-task networks} (fine-tuned for one task only) and add cross-stitch units? Or should one start with networks that have not been fine-tuned for the tasks? We explore the effect of both choices by initializing using two \emph{one-task networks} and two networks trained on ImageNet~\cite{imagenet,ILSVRC15}. We train the \emph{one-task} initialized cross-stitched network for 10k iterations and the ImageNet initialized cross-stitched network for 30k iterations (to account for the 20k fine-tuning iterations of the \emph{one-task} networks), and report the results in Table~\ref{tbl:imnet-init}. Task-specific initialization performs better than ImageNet initialization for both the tasks, which suggests that cross-stitching should be used after training task-specific networks.

\begin{table}[t]
\setlength{\tabcolsep}{0.2em}
\centering
\footnotesize{
\caption{We initialize the networks $\mathrm{A}$, $\mathrm{B}$ (from Figure~\ref{fig:cross-stitch-two}) from ImageNet, as well as task-specific networks. We observe that task-based initialization performs better than task-agnostic ImageNet initialization.}
\label{tbl:imnet-init}
\resizebox{\linewidth}{!}{
\begin{tabular}{@{}L{1.2cm} r*{7}{C{0.75cm}}@{}}
\toprule
& \multicolumn{5}{c}{{ Surface Normal}} & \multicolumn{3}{c}{{ Segmentation}} \\
\arrayrulecolor{Gray}
\cmidrule(l{-0.1cm}r{1.5pt}){2-6}
\cmidrule(l{1.5pt}){7-9}
\arrayrulecolor{black}
& \multicolumn{2}{>{\centering\let\newline\\\arraybackslash\hspace{-0.15cm}}m{1.59cm}}{Angle Distance} & \multicolumn{3}{c}{Within $t^{\circ} $} & & & \\

& \multicolumn{2}{>{\centering\let\newline\\\arraybackslash\hspace{-0.15cm}}m{1.59cm}}{(Lower Better)} & \multicolumn{3}{c}{(Higher Better)} & \multicolumn{3}{c}{(Higher Better)}\\
Init.\ & Mean & Med. & 11.25 & 22.5 & 30 & pixacc & mIU & fwIU \\
\midrule
ImageNet & 34.6	& 18.8 & 38.6& 53.7 & 59.4 & \textbf{48.0} & 17.7 & 33.4\\
One-task & \textbf{34.1} & \textbf{18.2} & \textbf{39.0} & \textbf{54.4} & \textbf{60.2} & 47.2 & \textbf{19.3} & \textbf{34.0} \\
\bottomrule
\end{tabular}
}
}
\end{table}

\subsection{Visualization of learned combinations}
\label{sec:viz-lin}
We visualize the weights $\alpha_\mathrm{S}$ and $\alpha_\mathrm{D}$ of the cross-stitch units for different initializations in Figure~\ref{tbl:lr-init-viz}. For this experiment, we initialize sub-networks $\mathrm{A}$ and $\mathrm{B}$ using \emph{one-task} networks and trained the cross-stitched network till convergence. Each plot shows (in sorted order) the $\alpha$ values for all the cross-stitch units in a layer (one per channel). We show plots for three layers: \texttt{pool1}, \texttt{pool5} and \texttt{fc7}. The initialization of cross-stitch units biases the network to start its training preferring a certain type of shared representation, \eg, $\left(\alpha_\mathrm{S},\alpha_\mathrm{D}\right)=(0.9,0.1)$ biases the network to learn more task-specific features, while $(0.5,0.5)$ biases it to share representations. Figure~\ref{tbl:lr-init-viz} (second row) shows that both the tasks, across all initializations, prefer a more task-specific representation for \texttt{pool5}, as shown by higher values of $\alpha_\mathrm{S}$. This is inline with the observation from Section~\ref{sec:intro-study} that Split \texttt{conv4} performs best for these two tasks. We also notice that the surface normal task prefers shared representations as can be seen by Figure~\ref{tbl:lr-init-viz}(b), where $\alpha_\mathrm{S}$ and $\alpha_\mathrm{D}$ values are in similar range.

\begin{table*}
\setlength{\tabcolsep}{0.1em}
\centering
\caption{We show the sorted $\alpha$ values (increasing left to right) for three layers. A higher value of $\alpha_\mathrm{S}$ indicates a strong preference towards task specific features, and a higher $\alpha_\mathrm{D}$ implies preference for shared representations. More detailed analysis in Section~\ref{sec:viz-lin}. Note that both $\alpha_\mathrm{S}$ and $\alpha_\mathrm{D}$ are sorted independently, so the channel-index across them do not correspond.}
\label{tbl:lr-init-viz}
\hspace*{-0.1in}
\begin{tabular}{@{}cccccccccccc@{}}
\toprule
Layer & \multicolumn{2}{c}{(a) $\alpha_{\mathrm{S}}=0.9, \alpha_{\mathrm{D}}=0.1$} & & \multicolumn{2}{c}{(b) $\alpha_{\mathrm{S}}=0.5, \alpha_{\mathrm{D}}=0.5$} & & \multicolumn{2}{c}{(c) $\alpha_{\mathrm{S}}=0.1, \alpha_{\mathrm{D}}=0.9$} \\
\midrule
& \hspace{0.1in} Segmentation & \hspace{0.1in} Surface Normal &  & \hspace{0.1in} Segmentation & \hspace{0.1in} Surface Normal & & \hspace{0.1in} Segmentation & \hspace{0.1in} Surface Normal \\

\texttt{pool1} &  \raisebox{-.5\height}{\includegraphics[width=0.16\textwidth]{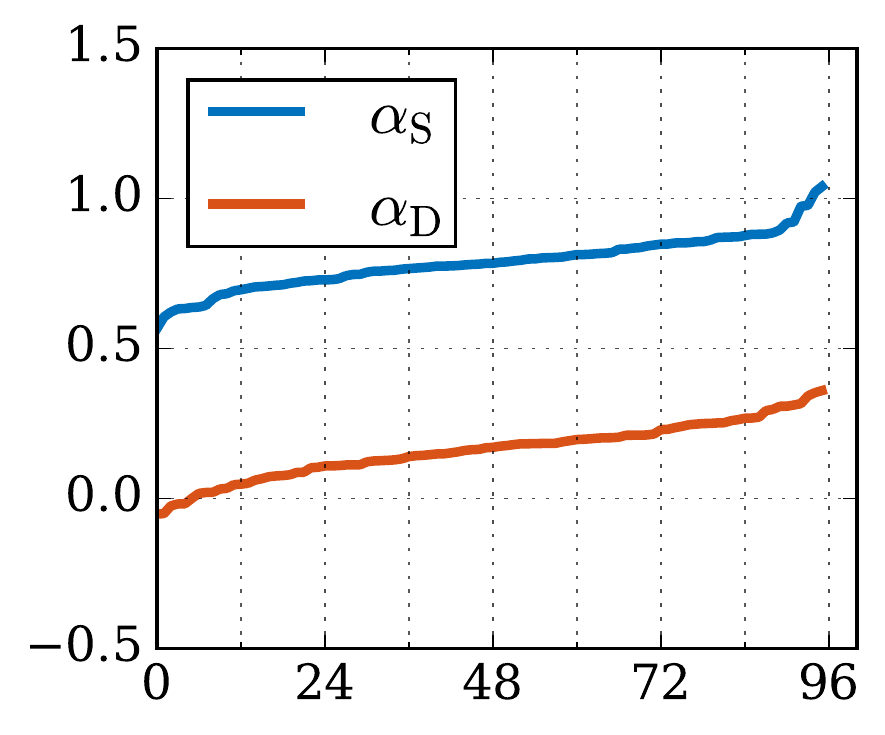}}
& \raisebox{-.5\height}{\includegraphics[width=0.16\textwidth]{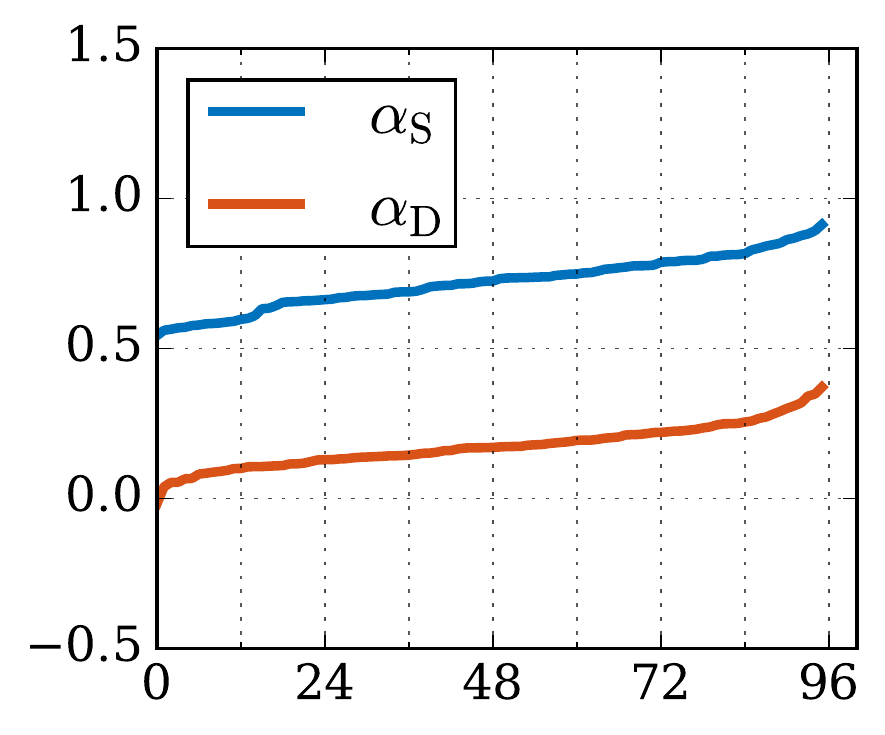}}
& &
\raisebox{-.5\height}{\includegraphics[width=0.16\textwidth]{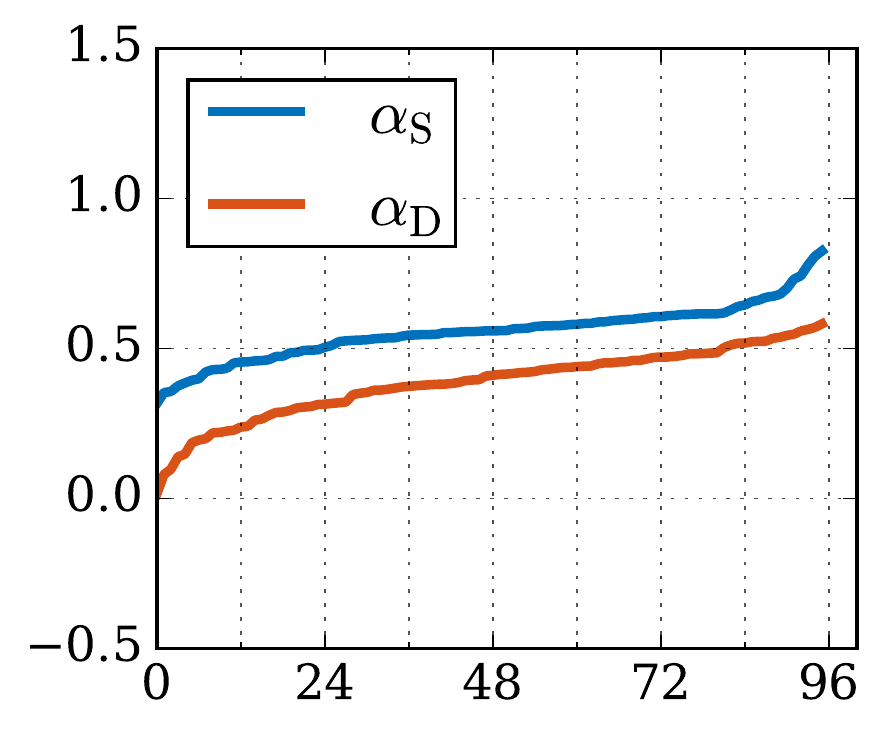}}
& \raisebox{-.5\height}{\includegraphics[width=0.16\textwidth]{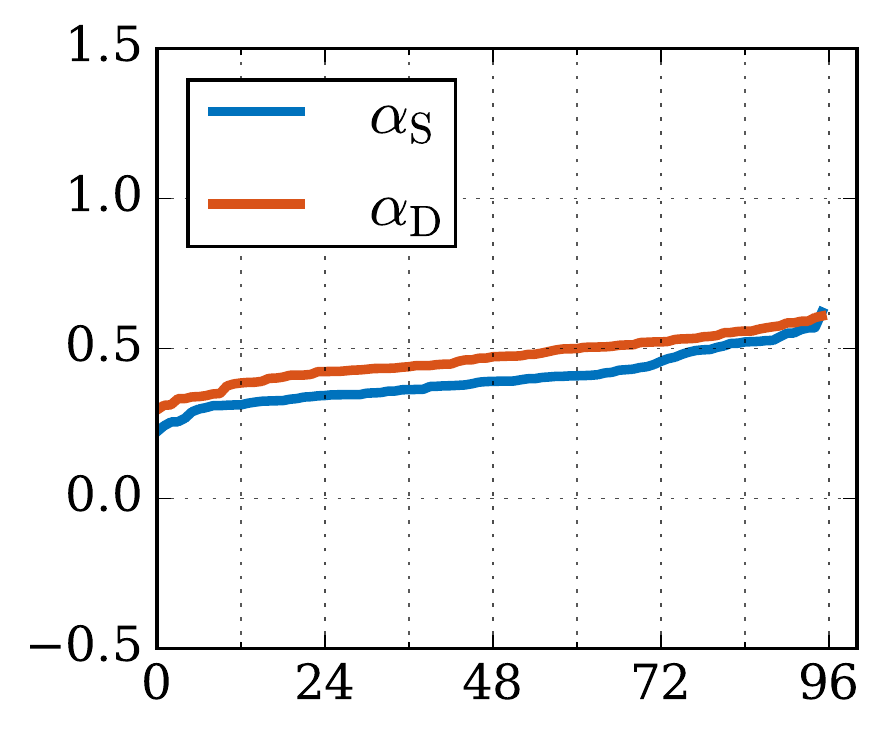}}
& &
\raisebox{-.5\height}{\includegraphics[width=0.16\textwidth]{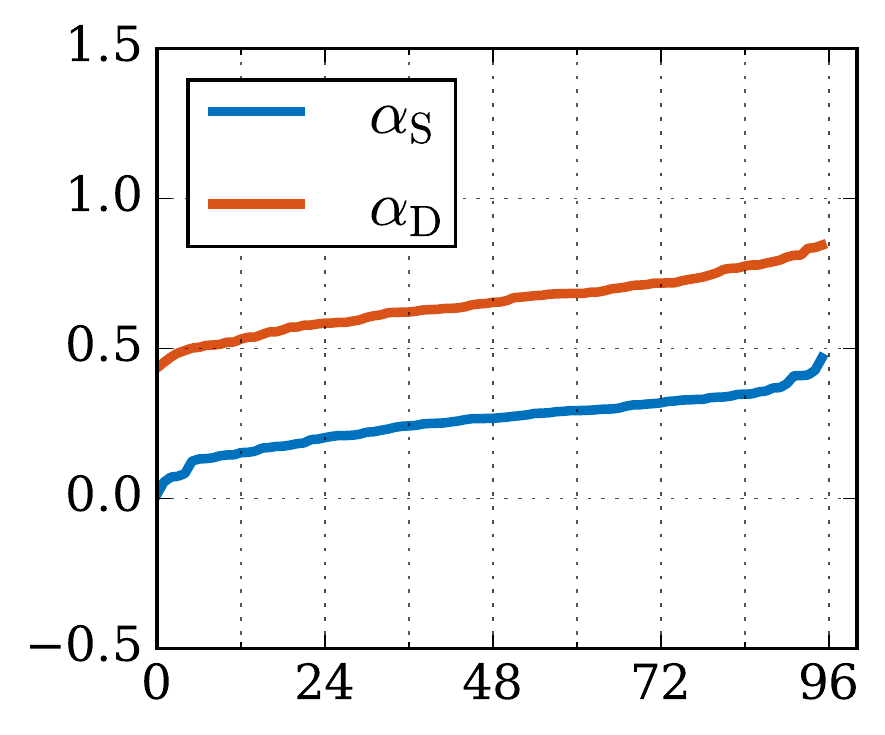}}
& \raisebox{-.5\height}{\includegraphics[width=0.16\textwidth]{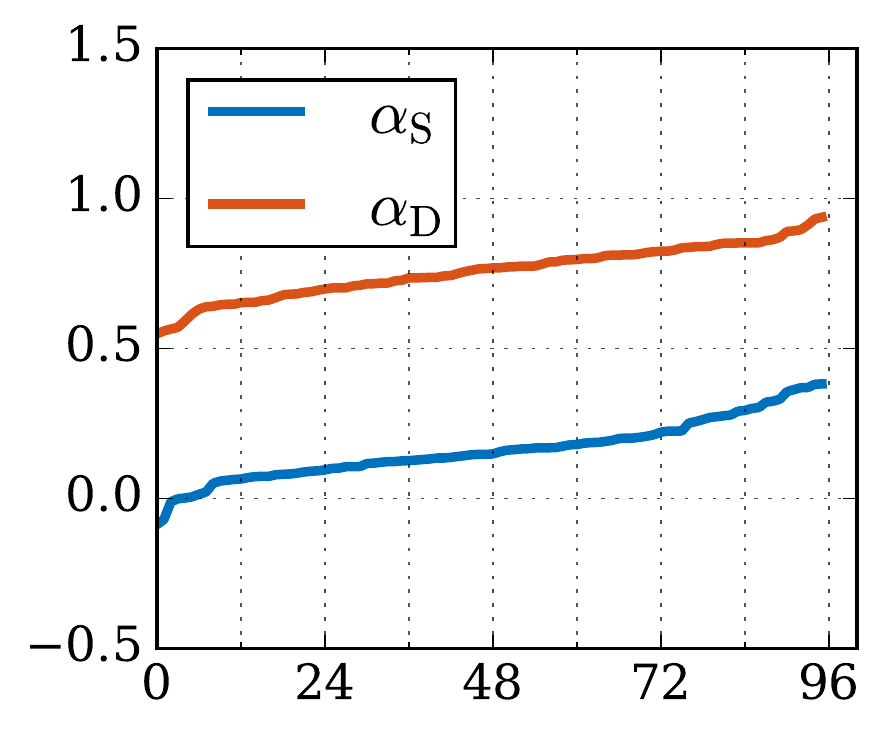}}
\\
\texttt{pool5} &  \raisebox{-.5\height}{\includegraphics[width=0.16\textwidth]{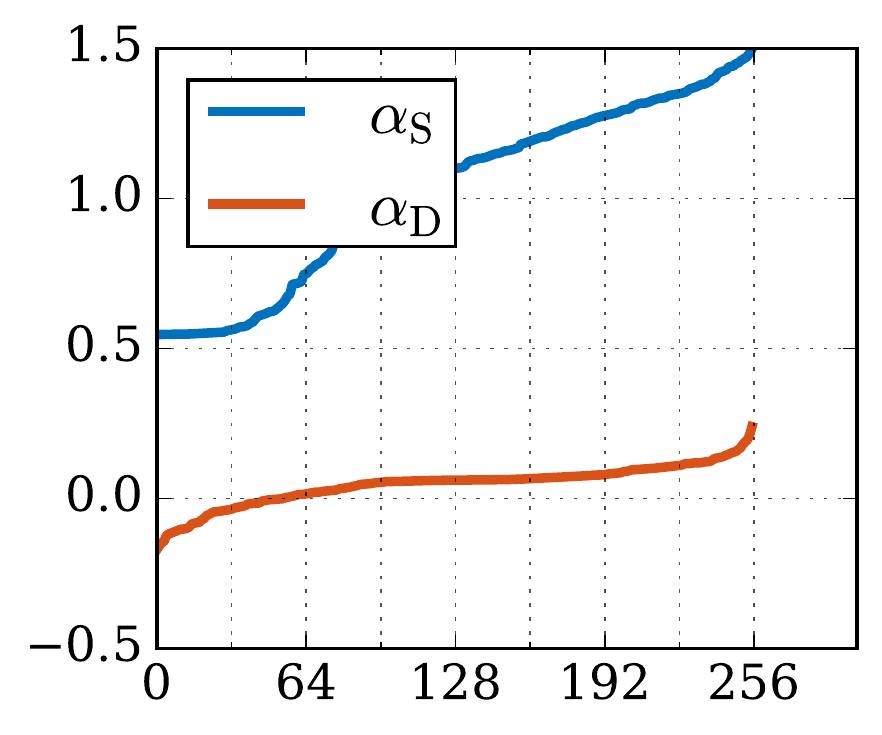}}
& \raisebox{-.5\height}{\includegraphics[width=0.16\textwidth]{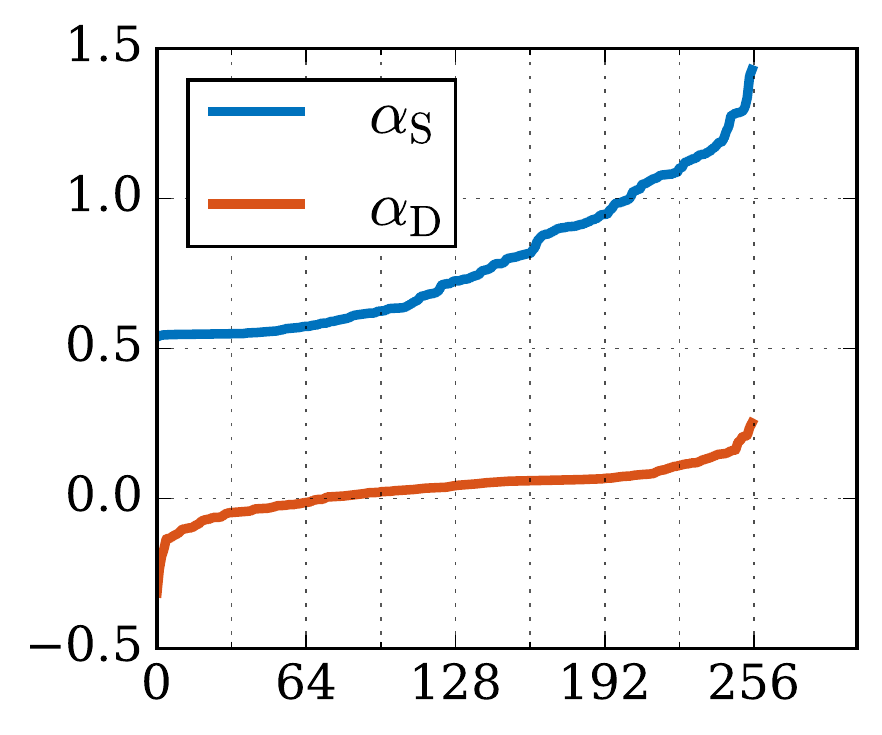}}
& &
\raisebox{-.5\height}{\includegraphics[width=0.16\textwidth]{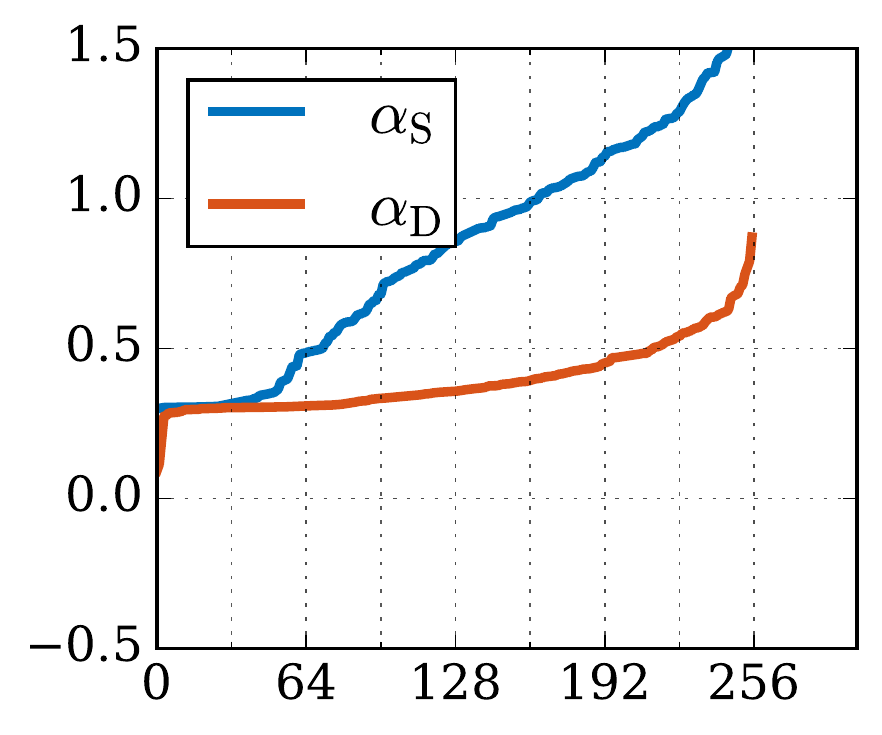}}
& \raisebox{-.5\height}{\includegraphics[width=0.16\textwidth]{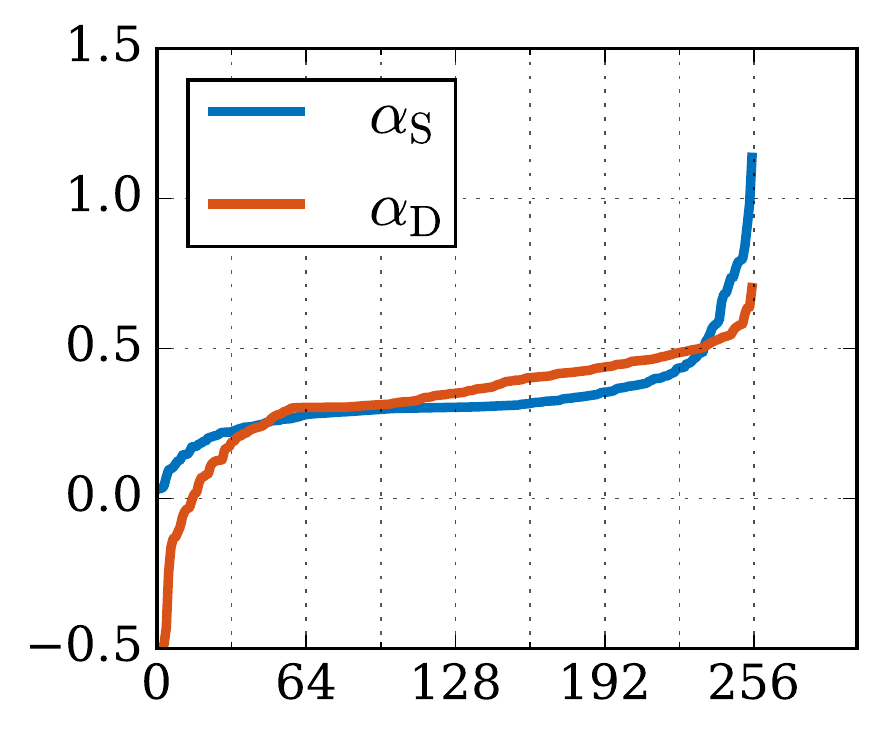}}
& &
\raisebox{-.5\height}{\includegraphics[width=0.16\textwidth]{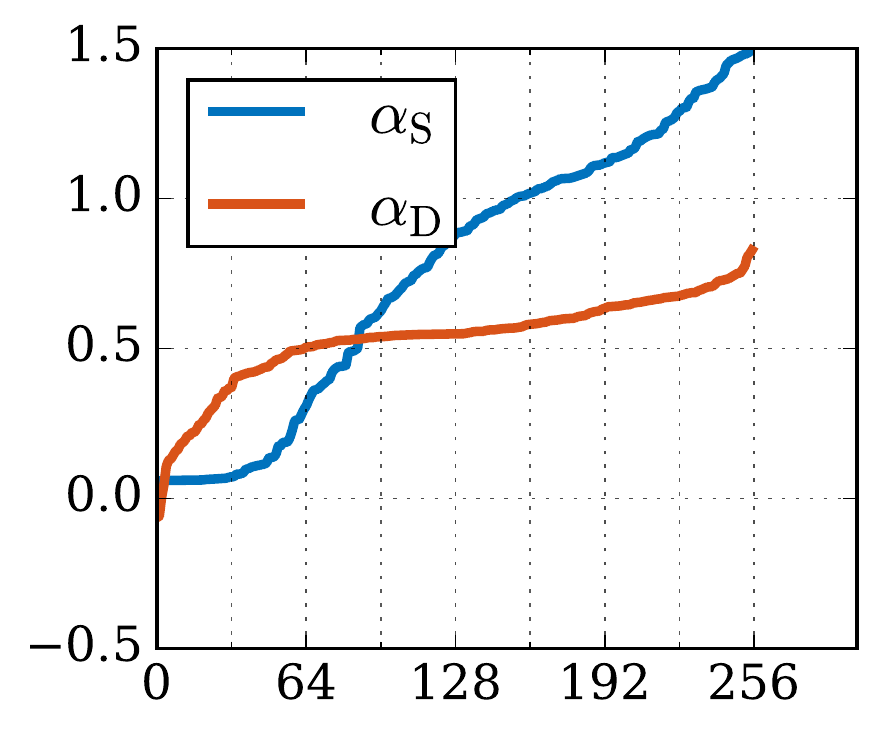}}
& \raisebox{-.5\height}{\includegraphics[width=0.16\textwidth]{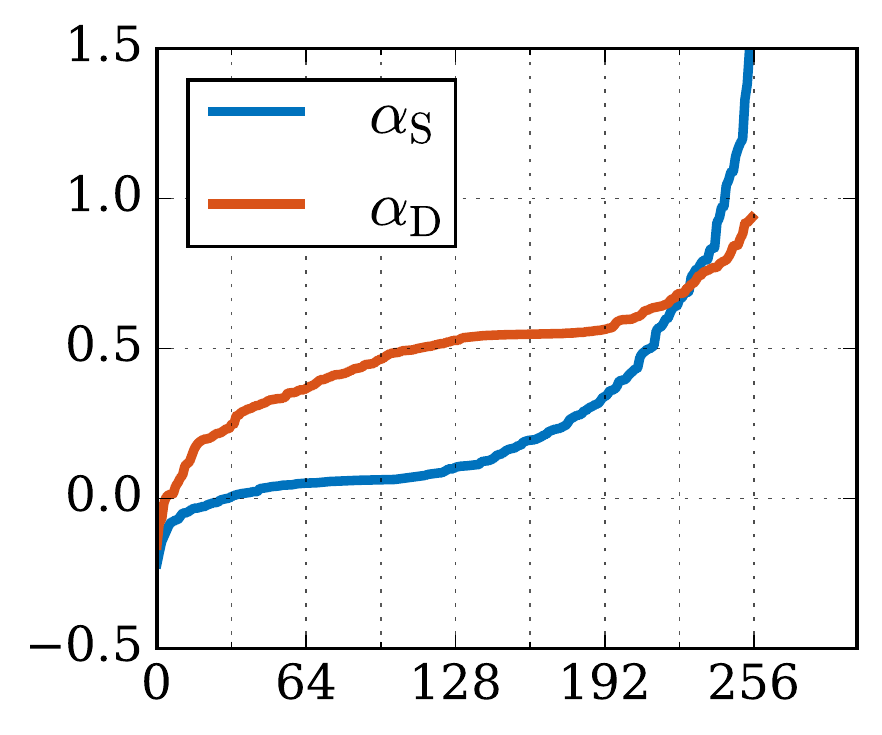}}
\\
\texttt{fc7} &  \raisebox{-.5\height}{\includegraphics[width=0.16\textwidth]{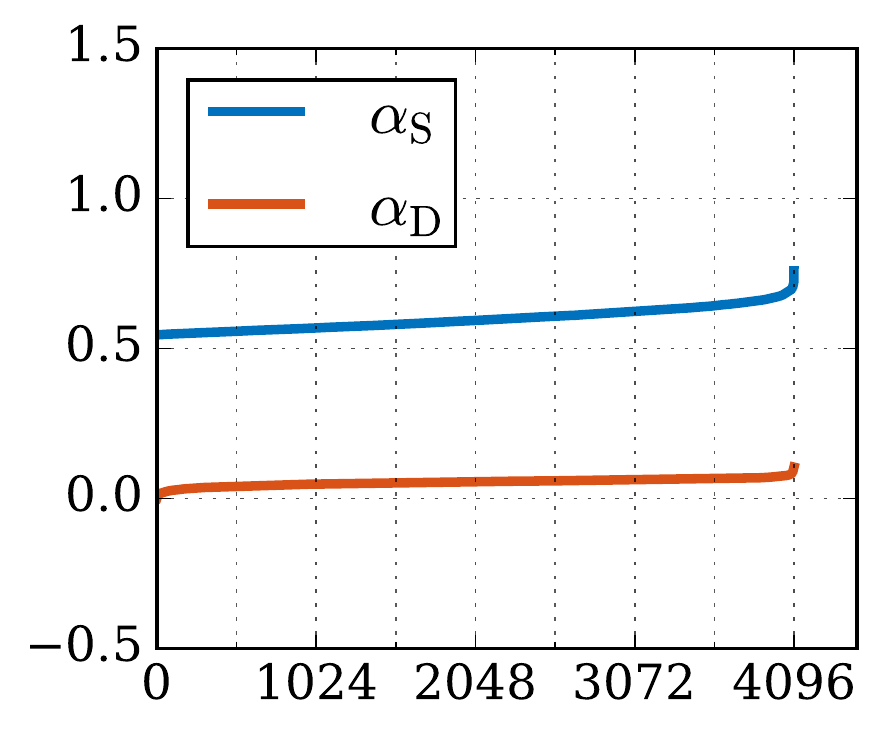}}
& \raisebox{-.5\height}{\includegraphics[width=0.16\textwidth]{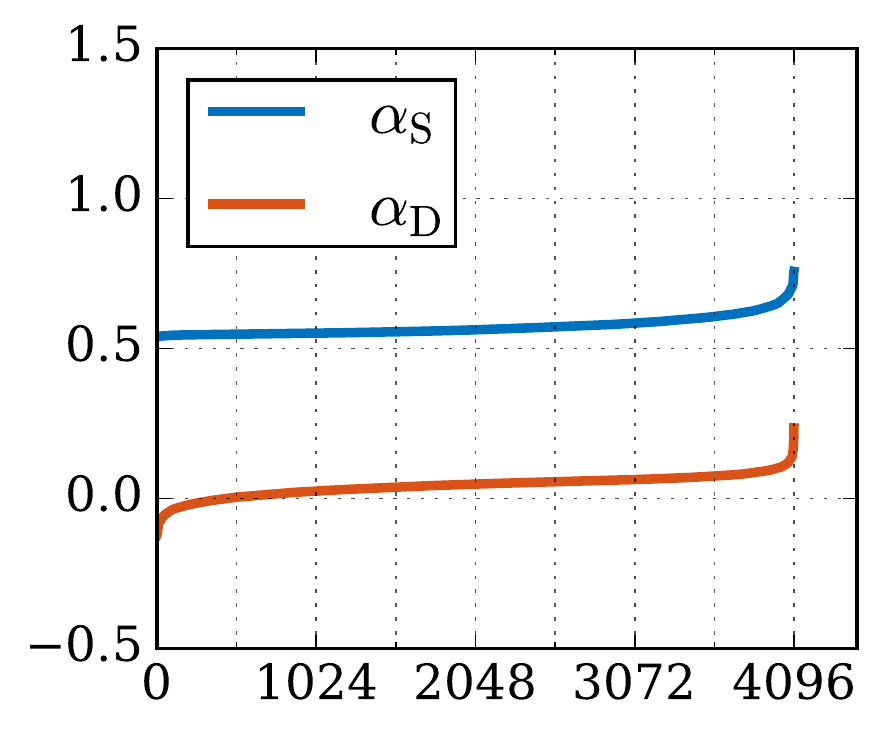}}
& &
\raisebox{-.5\height}{\includegraphics[width=0.16\textwidth]{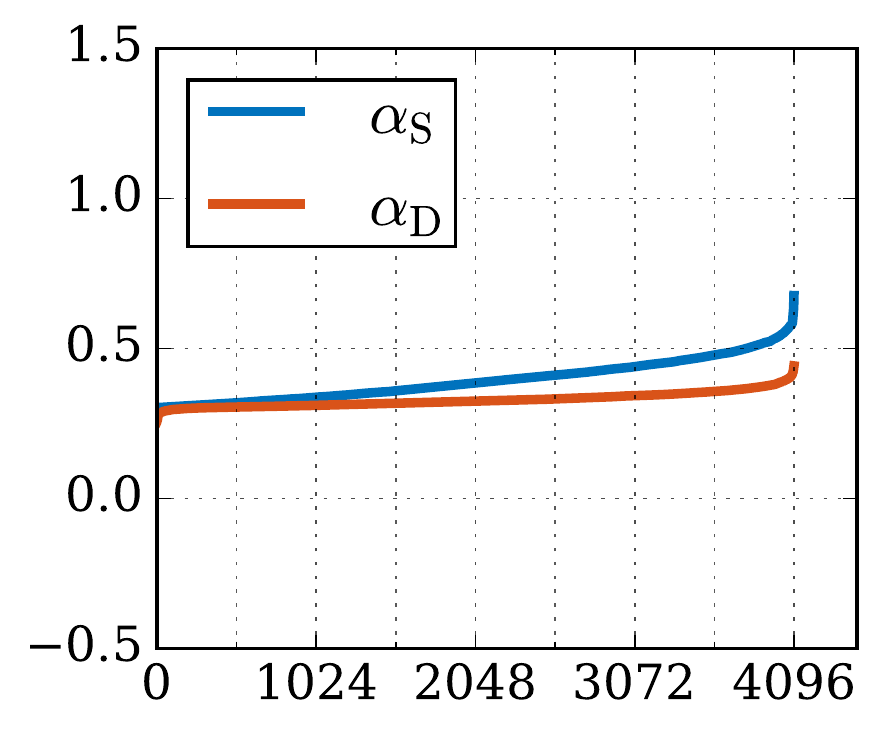}}
& \raisebox{-.5\height}{\includegraphics[width=0.16\textwidth]{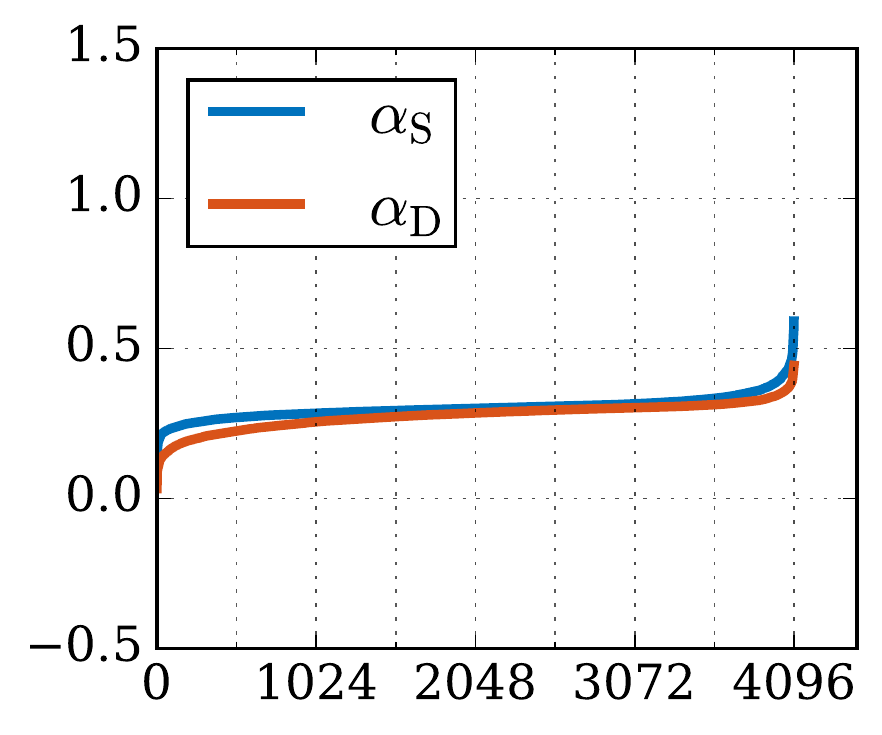}}
& &
\raisebox{-.5\height}{\includegraphics[width=0.16\textwidth]{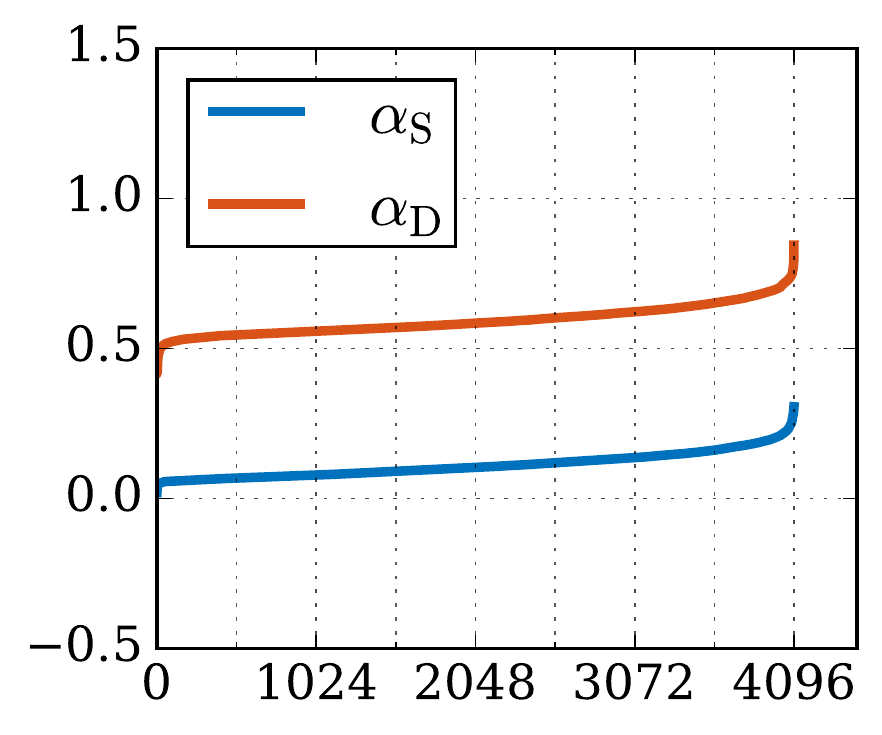}}
& \raisebox{-.5\height}{\includegraphics[width=0.16\textwidth]{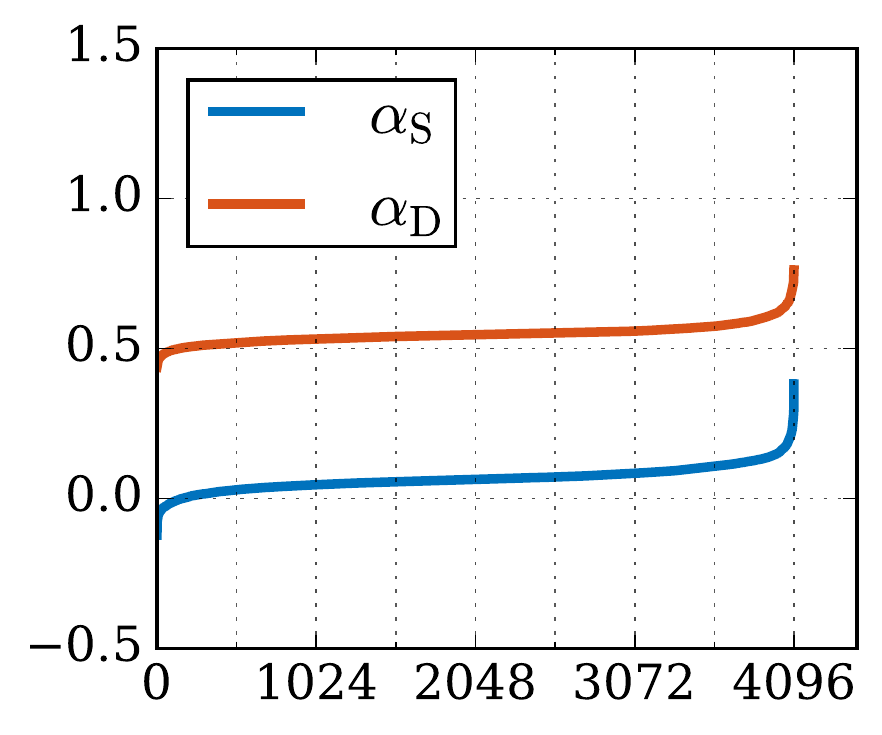}}
\\
\bottomrule
\end{tabular}
\end{table*}

\section{Experiments}
\label{sec:experiments}
\begin{figure*}[!ht]
\centering
\includegraphics[width=0.9\textwidth]{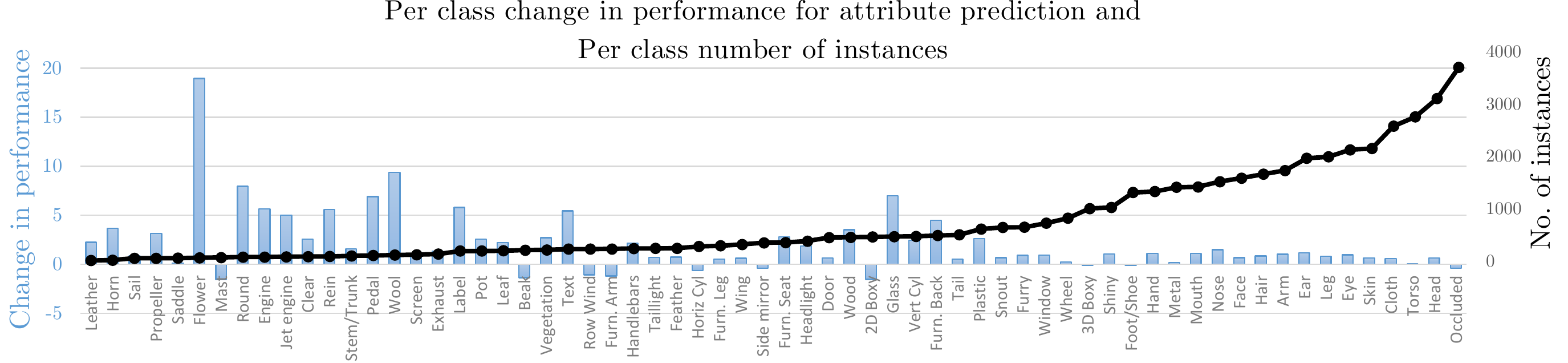}
\caption{Change in performance for attribute categories over the baseline is indicated by blue bars. We sort the categories in increasing order (from left to right) by the number of instance labels in the train set, and indicate the number of instance labels by the solid black line. The performance gain for attributes with lesser data (towards the left) is considerably higher compared to the baseline. We also notice that the gain for categories with lots of data is smaller.}
\vspace{-0.2in}
\label{fig:attr-starved-gain}
\end{figure*}
We now present experiments with cross-stitch networks for two pairs of tasks: semantic segmentation and surface normal prediction on NYU-v2~\cite{nyuv2}, and object detection and attribute prediction on PASCAL VOC 2008~\cite{pascal,apascal}. We use the experimental setup from Section~\ref{sec:ablation} for semantic segmentation and surface normal prediction, and describe the setup for detection and attribute prediction below.

\par \noindent \textbf{Dataset, Metrics and Network:} We consider the PASCAL VOC 20 classes for object detection, and the 64 attribute categories data from~\cite{apascal}. We use the PASCAL VOC 2008~\cite{pascal,apascal} dataset for our experiments and report results using the standard Average Precision (AP) metric. We start with the recent Fast-RCNN~\cite{fast-rcnn} method for object detection using the AlexNet~\cite{alexnet} architecture.

\par \noindent \textbf{Training:} For object detection, Fast-RCNN is trained using 21-way 1-vs-all classification with 20 foreground and 1 background class. However, there is a severe data imbalance in the foreground and background data points (boxes). To circumvent this, Fast-RCNN carefully constructs mini-batches with $1:3$ foreground-to-background ratio, \ie, at most 25\% of foreground samples in a mini-batch. Attribute prediction, on the other hand, is a multi-label classification problem with 64 attributes, which only train using foreground bounding boxes. To implement both tasks in the Fast R-CNN framework, we use the same mini-batch sampling strategy; and in every mini-batch only the foreground samples contribute to the attribute loss (and background samples are ignored).

\par \noindent \textbf{Scaling losses:} Both SemSeg and SN used same classification loss for training, and hence we were set their loss weights to be equal ($=1$). However, since object detection is formulated as 1-vs-all classification and attribute classification as multi-label classification, we balance the losses by scaling the attribute loss by $1/64$.

\par \noindent \textbf{Cross-stitching:} We combine two AlexNet architectures using the cross-stitch units after every pooling layer as shown in Figure~\ref{fig:cross-stitch-two}. In the case of object detection and attribute prediction, we use one cross-stitch unit per layer activation map. We found that maintaining a unit per channel, like in the case of semantic segmentation, led to unstable learning for these tasks. 

\subsection{Baselines}
We compare against four strong baselines for the two pairs of tasks and report the results in Table~\ref{tbl:sn-exp} and~\ref{tbl:obj-exp}.

\par \noindent \textbf{Single-task Baselines}: These serve as baselines without benefits of multi-task learning. First we evaluate a single network trained on only one task (denoted by `One-task') as described in Section~\ref{sec:ablation}. Since our approach cross-stitches two networks and therefore uses $2\times$ parameters, we also consider an ensemble of two one-task networks (denoted by `Ensemble'). However, note that the ensemble has $2\times$ network parameters for only one task, while the cross-stitch network has roughly $2\times$ parameters for two tasks. So for a pair of tasks, the ensemble baseline uses $\sim2\times$ the cross-stitch parameters.

\par \noindent \textbf{Multi-task Baselines}: The cross-stitch units enable the network to pick an optimal combination of shared and task-specific representation. We demonstrate that these units remove the need for finding such a combination by exhaustive brute-force search (from Section~\ref{sec:intro-study}). So as a baseline, we train all possible ``Split architectures'' for each pair of tasks and report numbers for the best Split for each pair of tasks.

There has been extensive work in Multi-task learning outside of the computer vision and deep learning community. However, most of such work, with publicly available code, formulates multi-task learning in an optimization framework that requires all data points in memory~\cite{zhou2012mutal,chen2011integrating,gu2009learning,zhou2012modeling,evgeniou2004regularized,lapin2014cvpr,su2015multi}. Such requirement is not practical for the vision tasks we consider.

So as our final baseline, we compare to a variant of~\cite{zhou2013learning,abdulnabi2015multi} by adapting their method to our setting and report this as `MTL-shared'. The original method treats each category as a separate `task', \emph{a separate network} is required for each category and \emph{all these networks are trained jointly}. Directly applied to our setting, this would require training 100s of ConvNets jointly, which is impractical. 
Thus, instead of treating each category as an independent task, we adapt their method to our two-task setting. We train these two networks jointly, using end-to-end learning, as opposed to their dual optimization to reduce hyperparameter search.

\subsection{Semantic Segmentation and Surface Normal Prediction}
Table~\ref{tbl:sn-exp} shows the results for semantic segmentation and surface normal prediction on the NYUv2 dataset~\cite{nyuv2}. We compare against two one-task networks, an ensemble of two networks, and the best Split architecture (found using brute force enumeration). The sub-networks $\mathrm{A}$, $\mathrm{B}$ (Figure~\ref{fig:cross-stitch-two}) in our cross-stitched network are initialized from the one-task networks. We use cross-stitch units after every pooling layer and fully connected layer (one per channel). Our proposed cross-stitched network improves results over the baseline one-task networks and the ensemble. Note that even though the ensemble has $2\times$ parameters compared to cross-stitched network, the latter performs better. Finally, our performance is better than the best Split architecture network found using brute force search. This shows that the cross-stitch units can effectively search for optimal amount of sharing in multi-task networks.

\begin{table}[t]
\setlength{\tabcolsep}{0.2em}
\centering
\footnotesize{
\caption{Surface normal prediction and semantic segmentation results on the NYU-v2~\cite{nyuv2} dataset. Our method outperforms the baselines for both the tasks.}
\label{tbl:sn-exp}
\resizebox{\linewidth}{!}{
\begin{tabular}{@{}L{2.2cm} r*{7}{C{0.75cm}}@{}}
\toprule
& \multicolumn{5}{c}{{ Surface Normal}} & \multicolumn{3}{c}{{ Segmentation}} \\
\arrayrulecolor{Gray}
\cmidrule(l{-0.1cm}r{1.5pt}){2-6}
\cmidrule(l{1.5pt}){7-9}
\arrayrulecolor{black}
& \multicolumn{2}{>{\centering\let\newline\\\arraybackslash\hspace{-0.15cm}}m{1.59cm}}{Angle Distance} & \multicolumn{3}{c}{Within $t^{\circ} $} & & & \\

& \multicolumn{2}{>{\centering\let\newline\\\arraybackslash\hspace{-0.15cm}}m{1.59cm}}{(Lower Better)} & \multicolumn{3}{c}{(Higher Better)} & \multicolumn{3}{c}{(Higher Better)}\\
Method & Mean & Med. & 11.25 & 22.5 & 30 & pixacc & mIU & fwIU \\
\midrule
\multirow{2}{*}{One-task}& 34.8 & 19.0 & 38.3	& 53.5 & 59.2 &- &- &- \\
& - & - & - & -  & -  & 46.6	& 18.4 & 33.1 \\
\arrayrulecolor{Gray}
\midrule
\multirow{2}{*}{Ensemble} & 34.4 & 18.5 & 38.7	& 54.2 & 59.7 &- &- &- \\
& - & - &- &-  & - & \textbf{48.2} & 18.9 & 33.8 \\
\midrule
Split \texttt{conv4} & 34.7 & 19.1 & 38.2 & 53.4 & 59.2 & 47.8 & 19.2 & 33.8 \\
\midrule
MTL-shared & 34.7 & 18.9 & 37.7 & 53.5 & 58.8 & 45.9 & 16.6 & 30.1 \\
\midrule
\arrayrulecolor{black}
Cross-stitch [\textbf{ours}] & \textbf{34.1} & \textbf{18.2} & \textbf{39.0} & \textbf{54.4} & \textbf{60.2} & 47.2 & \textbf{19.3} & \textbf{34.0} \\
\bottomrule
\end{tabular}
}
}
\end{table}

\subsection{Data-starved categories for segmentation}
\label{sec:seg-starved}
Multiple tasks are particularly helpful in regularizing the learning of shared representations\cite{teterwakshared,evgeniou2004regularized,caruanaThesis}. This regularization manifests itself empirically in the improvement of ``data-starved'' (few examples) categories and tasks. 

\begin{figure}
\includegraphics[width=0.45\textwidth]{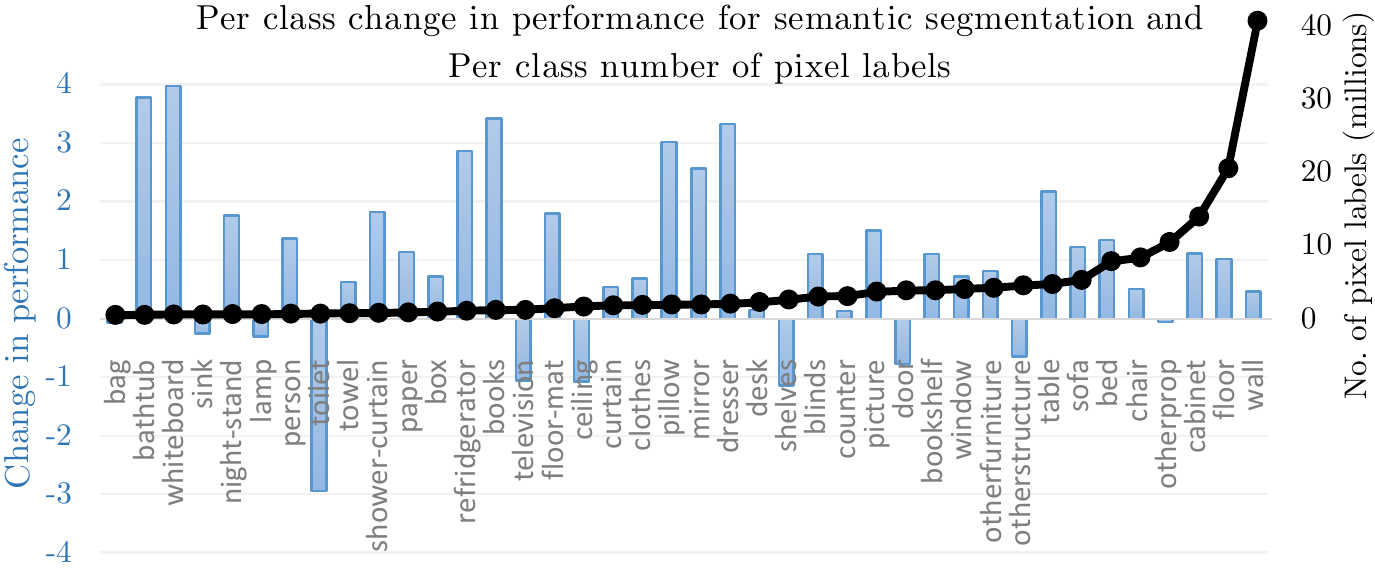}
\caption{Change in performance (meanIU metric) for semantic segmentation categories over the baseline is indicated by blue bars. We sort the categories (in increasing order from left to right) by the number of pixel labels in the train set, and indicate the number of pixel labels by a solid black line. The performance gain for categories with lesser data (towards the left) is more when compared to the baseline one-task network.}
\vspace{-0.1in}
\label{fig:data-starved-gain}
\end{figure}

For semantic segmentation, there is a high mismatch in the number of labels per category (see the black line in Figure~\ref{fig:data-starved-gain}). Some classes like \emph{wall}, \emph{floor} have many more instances than other classes like \emph{bag}, \emph{whiteboard} \etc. Figure~\ref{fig:data-starved-gain} also shows the per-class gain in performance using our method over the baseline one-task network. We see that cross-stitch units considerably improve the performance of ``data-starved'' categories (\eg, \emph{bag}, \emph{whiteboard}).

\subsection{Object detection and attribute prediction}
We train a cross-stitch network for the tasks of object detection and attribute prediction. We compare against baseline one-task networks and the best split architectures per task (found after enumeration and search, Section~\ref{sec:intro-study}). Table~\ref{tbl:obj-exp} shows the results for object detection and attribute prediction on PASCAL VOC 2008~\cite{pascal,apascal}. Our method shows improvements over the baseline for attribute prediction. It is worth noting that because we use a background class for detection, and not attributes (described in `Scaling losses' in Section~\ref{sec:experiments}), detection has many more data points than attribute classification (only 25\% of a mini-batch has attribute labels). Thus, we see an improvement for the data-starved task of attribute prediction. It is also interesting to note that the detection task prefers a shared representation (best performance by Split \texttt{fc7}), whereas the attribute task prefers a task-specific network (best performance by Split \texttt{conv2}).

\begin{table}[t]
\centering
\setlength{\tabcolsep}{0.2em}
\caption{Object detection and attribute prediction results on the attribute PASCAL~\cite{apascal} 2008 dataset}
\label{tbl:obj-exp}
{\small
\resizebox{\linewidth}{!}{
\begin{tabular}{{@{}L{3cm} C{2.5cm} C{2.5cm}@{}}}
\toprule
Method & Detection (mAP) & Attributes (mAP) \\
\midrule
\multirow{2}{*}{One-task} & 44.9 & - \\
& - & 60.9 \\
\arrayrulecolor{Gray}
\midrule
\multirow{2}{*}{Ensemble} & \textbf{46.1} & - \\
& - & 61.1 \\
\midrule
Split \texttt{conv2} & 44.6 & 61.0 \\
Split \texttt{fc7} & 44.8 & 59.7 \\
\midrule
MTL-shared & 42.7  & 54.1\\
\midrule
Cross-stitch [\textbf{ours}] & 45.2 & \textbf{63.0}\\
\arrayrulecolor{black}
\bottomrule
\end{tabular}
}
\vspace{-0.1in}
}
\end{table}

\subsection{Data-starved categories for attribute prediction}
Following a similar analysis to Section~\ref{sec:seg-starved}, we plot the relative performance of our cross-stitch approach over the baseline one-task attribute prediction network in Figure~\ref{fig:attr-starved-gain}. The performance gain for attributes with smaller number of training examples is considerably large compared to the baseline (\textbf{4.6}\% and \textbf{4.3}\% mAP for the top 10 and 20 attributes with the least data respectively). This shows that our proposed cross-stitch method provides significant gains for data-starved tasks by learning shared representations.

\section{Conclusion}
We present \emph{cross-stitch} units which are a generalized way of learning shared representations for multi-task learning in ConvNets. Cross-stitch units model shared representations as linear combinations, and can be learned end-to-end in a ConvNet. These units generalize across different types of tasks and eliminate the need to search through several multi-task network architectures on a per task basis. We show detailed ablative experiments to see effects of hyperparameters, initialization \etc when using these units. We also show considerable gains over the baseline methods for data-starved categories. Studying other properties of cross-stitch units, such as where in the network should they be used and how should their weights be constrained, is an interesting future direction.

{\small
\paragraph{Acknowledgments:} We would like to thank Alyosha Efros and Carl Doersch for helpful discussions. This work was supported in part by ONR MURI N000141612007 and the US Army Research Laboratory (ARL) under the CTA program (Agreement W911NF-10-2-0016). AS was supported by the MSR fellowship. We thank NVIDIA for donating GPUs.}

\pagebreak
{\small
\bibliographystyle{ieee}
\bibliography{mRefs}
}

\end{document}